\documentclass{ceurart}

\usepackage{graphicx}
\usepackage{booktabs}
\usepackage{float}
\usepackage{latexsym}
\usepackage{url}
\usepackage[T1]{fontenc}
\usepackage[utf8]{inputenc}
\usepackage[colorinlistoftodos]{todonotes}
\usepackage{tikz}
\usetikzlibrary{shapes.geometric, arrows.meta, positioning, calc}

\copyrightyear{2026}
\copyrightclause{Copyright for this paper by its authors. 
  Use permitted under Creative Commons License Attribution 4.0 International (CC BY 4.0).}
\conference{iTextbooks'26: Seventh Workshop on Intelligent Textbooks,
  June 28, 2026, Seoul, Republic of Korea}

\begin{document}

\title{Automated Textbook Auditing with Multi-Agent LLM Systems}

\author[1]{Ciprian Cristescu}[%
  email=cristescuciprian.cc@gmail.com,
]
\author[1]{Adrian-Marius Dumitran}[%
  email=marius.dumitran@unibuc.ro,
  orcid=0009-0005-3547-5772,
]
\author[2]{Angela-Liliana Dumitran}[%
  email=dumitranangela@gmail.com,
  orcid=0009-0003-3590-9441,
]
\author[1]{Gabriel Ștefan}[%
  email=gabrielstefan04@gmail.com,
  orcid=0009-0007-4193-7181,
]
\address[1]{University of Bucharest, Bucharest, Romania}
\address[2]{Dimitrie Cantemir Christian University, Bucharest, Romania}

\begin{abstract}
Ensuring the quality of educational materials requires more
than standard proofreading: textbooks must be audited for
factual accuracy, domain-specific technical correctness, and
linguistic quality simultaneously --- a task that general-purpose
grammar checkers cannot address. We present \textbf{AI Textbook
Auditor}, a modular multi-agent pipeline for automated quality
assurance of educational materials across subject domains.
The system accepts a textbook PDF and produces a structured,
human-reviewable report via two analysis tracks: a
\textbf{Factual and Technical Track} in which an ensemble of
specialized LLM agents detects factual inaccuracies, code
errors, incorrect definitions, and conceptual inconsistencies,
augmented with web search for humanities domains; and
a \textbf{Grammar Track} operating PDF-natively to preserve
diacritical encoding. A \textbf{Judge Agent} filters false
positives using domain-specific rules before presenting
findings to a human reviewer. The pipeline supports two
ingestion modes --- vision-native page rendering and
PyMuPDF text extraction --- and is domain-adaptable via
custom prompts encoding subject-specific error taxonomies.
We demonstrate the system on two Romanian upper-secondary
textbooks: a CS textbook (56 technical findings across seven
categories, with an expert-validated precision of 62.5\%) and
a history and social sciences textbook (72 findings spanning
factual errors, ideological bias, and grammar). The system is designed as
a triage tool that reduces the manual effort of locating
candidate issues, with human expert validation required
before any editorial action.
\end{abstract}

\maketitle
\section{Introduction}

High-quality textbooks are foundational to effective teaching,
yet ensuring their quality at scale remains an open challenge.
General-purpose grammar checkers address only the linguistic
dimension and are blind to the error types that matter most
in technical and scientific textbooks: a misplaced stream
operator in a C/C++ example, an extra mathematical symbol, a historical date that contradicts the
established record, or a mountain's altitude stated incorrectly
in a geography chapter.

We present \textbf{AI Textbook Auditor}, a modular multi-agent
system for automated quality assurance of educational materials.
An ensemble of \textbf{specialized LLM Agents} analyzes textbook content
for factual inaccuracies and domain-specific technical errors,
followed by a \textbf{Judge Agent} that filters false positives
before presenting findings to a human reviewer. The system is
domain-adaptable with the help of a domain expert via a custom prompt specifying the subject,
error taxonomy, and negative constraints; for humanities
textbooks it is augmented with web search for claim
verification, while for technical subjects it relies on
parametric model knowledge. We demonstrate the system on Romanian upper-secondary textbooks across two subject domains.

The main contributions of this work are:
\textbf{(i)} a modular multi-agent auditing pipeline with
domain adaptation via custom prompts;
\textbf{(ii)} a Judge Agent for domain-aware false-positive
filtering without additional human annotation in the detection
loop;
\textbf{(iii)} a unified architecture supporting web-search
mode for humanities and parametric-knowledge mode for technical
domains;
\textbf{(iv)} preliminary evaluation on two Romanian upper-secondary textbooks spanning CS and humanities domains.

\section{Related Work}

\textbf{Automated educational content assessment} has been studied 
primarily from the perspective of student-produced text: Grammar 
Error Correction \cite{ng2014conll, bryant2023grammatical}, automatic 
scoring \cite{burrows2015eras, attali2006automated}, and feedback 
generation \cite{kochmar2020automated}. Relatively little work targets the quality of published 
textbook content itself, and to our knowledge no prior 
system addresses the combined detection of factual 
inaccuracies and domain-specific technical errors across 
subject areas --- CS textbooks being one concrete instance 
of a broader quality assurance challenge

\textbf{Multi-agent LLM architectures} have demonstrated that role 
specialization and structured inter-agent communication improve 
performance on complex reasoning tasks \cite{hong2024metagpt}. Our 
system applies this principle to textbook auditing, combining a 
domain-adaptable technical agent ensemble with a dedicated Judge 
Agent for false-positive filtering.

\textbf{Code error detection} via static analysis and compiler-based 
tools works well for compilable code but cannot handle the incomplete 
fragments, pseudocode hybrids, and illustrative snippets typical of 
textbook content. LLMs have shown strong performance on code 
understanding and error detection tasks \cite{chen2021codex, 
roziere2023codellama, guo2024deepseekcoderlargelanguagemodel}, making them well-suited for this setting.

\textbf{Web-augmented fact-checking} with LLMs has been explored 
for claim verification tasks \cite{thorne2018fever, lewis2020rag}. 
Our system integrates optional web search for humanities domains 
where parametric model knowledge is insufficient, while relying on 
parametric knowledge alone for technical subjects where language 
standards and algorithmic theory provide a stable verification basis.

\section{System Architecture}

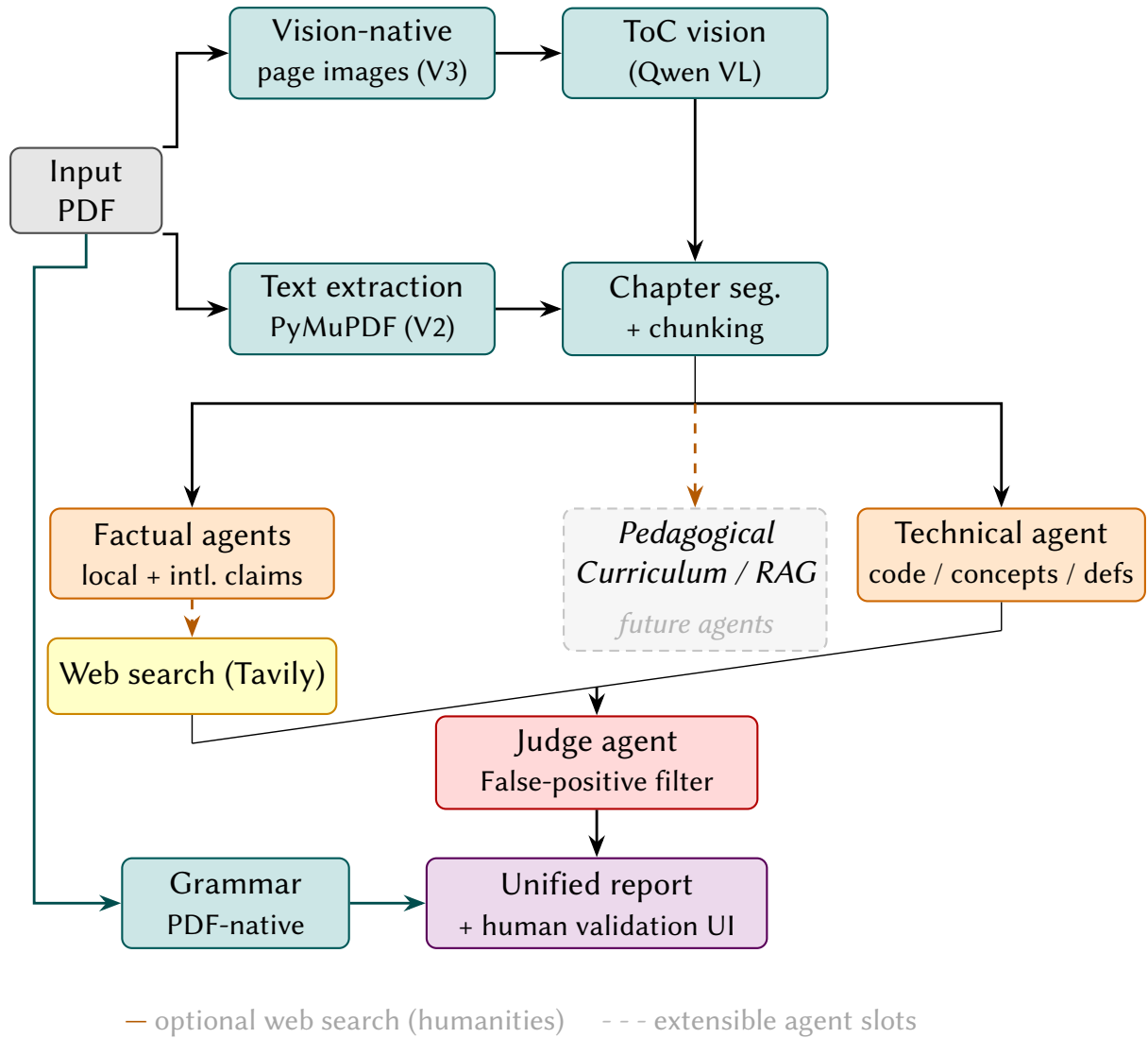
\begin{figure*}[ht]
\centering
\resizebox{\linewidth}{!}{%
\begin{tikzpicture}[
    node distance=0.4cm and 0.5cm,
    pdfbox/.style={
        rectangle, rounded corners=3pt, line width=0.6pt,
        draw=gray!60!black, fill=gray!20,
        minimum height=0.9cm, minimum width=1.6cm,
        font=\small, align=center},
    ingestbox/.style={
        rectangle, rounded corners=3pt, line width=0.6pt,
        draw=teal!70!black, fill=teal!20,
        minimum height=0.95cm, minimum width=2.8cm,
        font=\small, align=center},
    agentbox/.style={
        rectangle, rounded corners=3pt, line width=0.6pt,
        draw=orange!80!black, fill=orange!20,
        minimum height=0.95cm, minimum width=3.0cm,
        font=\small, align=center},
    webbox/.style={
        rectangle, rounded corners=3pt, line width=0.6pt,
        draw=orange!70!yellow!80!black, fill=yellow!22,
        minimum height=0.8cm, minimum width=3.0cm,
        font=\small, align=center},
    futurebox/.style={
        rectangle, rounded corners=3pt, line width=0.6pt,
        draw=gray!45, fill=gray!7,
        minimum height=1.3cm, minimum width=2.4cm,
        font=\small\itshape, align=center,
        densely dashed},
    judgebox/.style={
        rectangle, rounded corners=3pt, line width=0.6pt,
        draw=red!70!black, fill=red!15,
        minimum height=0.95cm, minimum width=3.4cm,
        font=\small, align=center},
    grammarbox/.style={
        rectangle, rounded corners=3pt, line width=0.6pt,
        draw=teal!70!black, fill=teal!20,
        minimum height=0.95cm, minimum width=2.4cm,
        font=\small, align=center},
    reportbox/.style={
        rectangle, rounded corners=3pt, line width=0.6pt,
        draw=violet!70!black, fill=violet!15,
        minimum height=0.95cm, minimum width=3.6cm,
        font=\small, align=center},
    arr/.style={-Stealth, thick},
    optarr/.style={-Stealth, thick, dashed, draw=orange!70!black},
    garr/.style={-Stealth, thick, draw=teal!60!black},
]

\node[pdfbox] (pdf) {Input\\PDF};

\node[ingestbox, above right=0.5cm and 0.7cm of pdf] (visionmode)
    {Vision-native\\{\footnotesize page images (V3)}};

\node[ingestbox, below right=0.3cm and 0.7cm of pdf] (textmode)
    {Text extraction\\{\footnotesize PyMuPDF (V2)}};

\draw[arr] (pdf.north east) -- ++(0.15,0) |- (visionmode.west);
\draw[arr] (pdf.south east) -- ++(0.15,0) |- (textmode.west);

\node[ingestbox, right=0.7cm of visionmode] (toc)
    {ToC vision\\{\footnotesize (Qwen VL)}};

\node[ingestbox, right=0.7cm of textmode] (chapseg)
    {Chapter seg.\\{\footnotesize + chunking}};

\draw[arr] (visionmode.east) -- (toc.west);
\draw[arr] (textmode.east)   -- (chapseg.west);
\draw[arr] (toc.south)       -- (chapseg.north);

\coordinate (junc) at ($(chapseg.south) + (0,-0.5)$);
\draw (chapseg.south) -- (junc);

\node[agentbox, below left=1.6cm and 2.4cm of chapseg] (factual)
    {Factual agents\\{\footnotesize local + intl.\ claims}};

\node[agentbox, below right=1.6cm and 0.3cm of chapseg] (technical)
    {Technical agent\\{\footnotesize code / concepts / defs}};

\draw[arr] (junc) -| (factual.north);
\draw[arr] (junc) -| (technical.north);

\node[futurebox, below=1.6cm of chapseg] (future)
    {Pedagogical\\Curriculum / RAG\\[3pt]{\footnotesize\color{gray!60} future agents}};

\draw[optarr] (junc) -- (future.north);

\node[webbox, below=0.4cm of factual] (websearch)
    {Web search (Tavily)};

\draw[optarr] (factual.south) -- (websearch.north);

\coordinate (collL) at ($(websearch.south) + (0,-0.3)$);
\coordinate (collR) at ($(technical.south) + (0,-0.3)$);
\coordinate (collM) at ($(collL)!0.5!(collR)$);

\draw (websearch.south) -- (collL);
\draw (technical.south) -- (collR);
\draw (collL)           -- (collR);

\node[judgebox, below=0.3cm of collM] (judge)
    {Judge agent\\{\footnotesize False-positive filter}};

\draw[arr] (collM) -- (judge.north);

\node[reportbox, below=0.5cm of judge] (report)
    {Unified report\\{\footnotesize + human validation UI}};

\draw[arr] (judge.south) -- (report.north);

\node[grammarbox, left=0.8cm of report] (grammar)
    {Grammar\\{\footnotesize PDF-native}};

\draw[garr] (pdf.south) -- ++(0,-0.35)
    -- ++(-0.55,0)
    |- (grammar.west);

\draw[garr] (grammar.east) -- (report.west);

\node[below=0.5cm of report, xshift=-0.8cm, font=\footnotesize,
      text=gray!70] (leg1)
    {\textcolor{orange!70!black}{---} optional web search (humanities)
     \quad
     \textcolor{gray!55}{- - -} extensible agent slots};

\end{tikzpicture}%
}
\caption{%
Overview of the auditing pipeline.
}
\label{fig:pipeline}
\end{figure*}

The pipeline accepts a textbook PDF and produces a structured,
human-reviewable report of candidate errors. Figure~\ref{fig:pipeline}
provides an overview of the end-to-end architecture. All components
are implemented as a Streamlit application with exportable CSV and
Excel output.

\subsection{Ingestion and Document Segmentation}



The system supports two ingestion modes: \textbf{vision-native}
(V3), where pages are rendered to images for a multimodal LLM,
avoiding the diacritical encoding artifacts common in Romanian
publisher PDFs; and \textbf{text-extraction} (V2), using PyMuPDF
with Unicode NFC normalization for lower cost on large documents.
Both share the same downstream agents and output schema. Document
structure is induced from the table of contents via a vision step
extracting \emph{(chapter title, start page)} pairs, editable
before analysis; ToC failure falls back to fixed-size page blocks.
Chapter text exceeding the context window is split with a
sliding-window chunker (9\,000 characters, 700-character overlap)
to avoid truncating code examples.

\subsection{Agent Ensemble}

The analytical core comprises two specialized agents operating
in parallel on each chapter chunk.

\paragraph{Factual agent.} The Factual Agent extracts verifiable
claims from the text and checks them for accuracy. For
\textbf{humanities domains} (history, geography, literature) it
is augmented with web search via Tavily, issuing targeted queries
per claim and grounding verdicts in retrieved evidence. For
\textbf{technical and exact-science domains} it relies on
parametric model knowledge, which is sufficient for verifying
code correctness, standard compliance, and algorithmic claims
without retrieval overhead.

\paragraph{Technical agent.} The Technical Agent supports two
prompt modes: a \textbf{generic} cross-domain mode that detects
factual, mathematical, scientific, conceptual, contradiction, and
code/logic errors without a fixed taxonomy, and a \textbf{custom}
mode in which a domain-specific error taxonomy is encoded directly
in the prompt. For computer science textbooks the custom mode
operates against an eight-category taxonomy
(Table~\ref{tab:taxonomy}) derived from empirical analysis of
Romanian CS textbooks, covering invalid syntax, logic errors,
pseudocode mistakes, incorrect concept definitions, non-standard
constructs, and portability issues; the taxonomy is replaceable
for other technical domains. Either mode is selectable per run.

\begin{table}[ht]
\centering
\caption{Technical error taxonomy for CS textbooks,
encoded as domain-specific prompt constraints.}
\label{tab:taxonomy}
\small
\begin{tabular}{@{}ll@{}}
\toprule
\textbf{Category} & \textbf{Description / example} \\
\midrule
\textsc{Syntax}      & Invalid C/C++ syntax: \texttt{int:x;}, \texttt{Typedef struct}, missing semicolons \\
\textsc{Code}        & Logic errors: \texttt{and} instead of \texttt{\&\&}, inconsistent loop variable, literal in condition \\
\textsc{Pseudocode}  & Digit 0 as letter O, wrong assignment operator, logic contradicts surrounding explanation \\
\textsc{Example}     & Code or pseudocode output contradicts the described result \\
\textsc{Concept}     & False language behavior claims: ``the first array element is \texttt{T[1]}'' in C/C++ \\
\textsc{Definition}  & Wrong data structure or algorithm definition: stack described as FIFO, incorrect complexity \\
\textsc{Standard}    & \texttt{void main()} or \texttt{\#include <iostream.h>} presented without qualification as current standard \\
\textsc{Portability} & Compiler-specific patterns presented as universally valid without caveat \\
\bottomrule
\end{tabular}
\end{table}

Both agents follow a consistent prompt engineering pattern:
role specialization, explicit negative constraints (what not to
report), strict JSON schema forcing, and an empty-list instruction
to suppress hallucinated findings when no clear error is present.

\paragraph{Extensibility.} The architecture accommodates
additional specialized agents without modifying the core
pipeline --- candidates include Pedagogical, Bias, Curriculum Alignment,
and RAG-based verification agents (shown as dashed slots
in Figure~\ref{fig:pipeline}).

\subsection{Judge Agent}

Raw agent outputs exhibit a non-trivial false-positive rate from
two systematic sources: encoding artifacts in extracted text
(corrupted diacritical characters misread as content errors) and
over-flagging of valid domain conventions (e.g.\ 1-based indexing
in Pascal contexts, standard Romanian CS terminology). The Judge
Agent performs a second-pass LLM call over aggregated findings,
processing them in batches and returning a binary verdict
(\texttt{validat: true/false}) with a brief justification. Its
prompt encodes explicit false-positive and true-positive criteria
derived from observed failure modes, requiring no additional
human annotation in the detection loop.

\subsection{Domain Adaptation via Custom Prompts}

The agent ensemble is fully parameterized by a \textbf{custom prompt}
specifying the subject domain, error taxonomy, and negative
constraints for the textbook under review. Authoring this prompt
requires a \textbf{domain expert}: the quality of the audit depends
directly on a subject specialist encoding the relevant error
categories and the conventions that must \emph{not} be flagged, since
a mis-specified taxonomy is the main source of false positives.
Given such a prompt, switching domains --- e.g.\ from CS to history
--- requires only replacing the taxonomy and enabling web search,
while the pipeline, Judge, and interface remain unchanged.

\subsection{Grammar Module}

A parallel grammar module, authored with a specialist in Romanian
orthography, checks linguistic compliance independently of the
factual track; its findings bypass the Judge Agent and default to
unvalidated pending review. As the norms live in the prompt rather
than the architecture, the module is adaptable to other languages.
Its full design and evaluation are a separate contribution beyond
this paper's scope; we note it here as a parallel track
(Figure~\ref{fig:pipeline}).

\subsection{Human Validation Interface}
All findings are consolidated into a unified report
(\texttt{validated, type, chapter, fragment, suggestion,
explanation}) presented as an editable data grid.
Reviewers toggle a \texttt{validated} checkbox per finding
and export validated results as CSV or Excel.
\section{Preliminary Evaluation}

We report illustrative findings from two pilot runs on
Romanian upper-secondary textbooks from different subject
domains, demonstrating the system across both implemented
agent configurations. 





\subsection{CS Domain: Upper-Secondary School Textbook}
The technical agent was run, in its generic cross-domain prompt
mode, on a 283-page Romanian CS upper-secondary textbook,
producing 56 candidate findings across seven categories, as
shown in Figure~\ref{fig:report}. Representative confirmed
findings include:

\begin{figure}[ht]
\centering
\includegraphics[width=\linewidth]{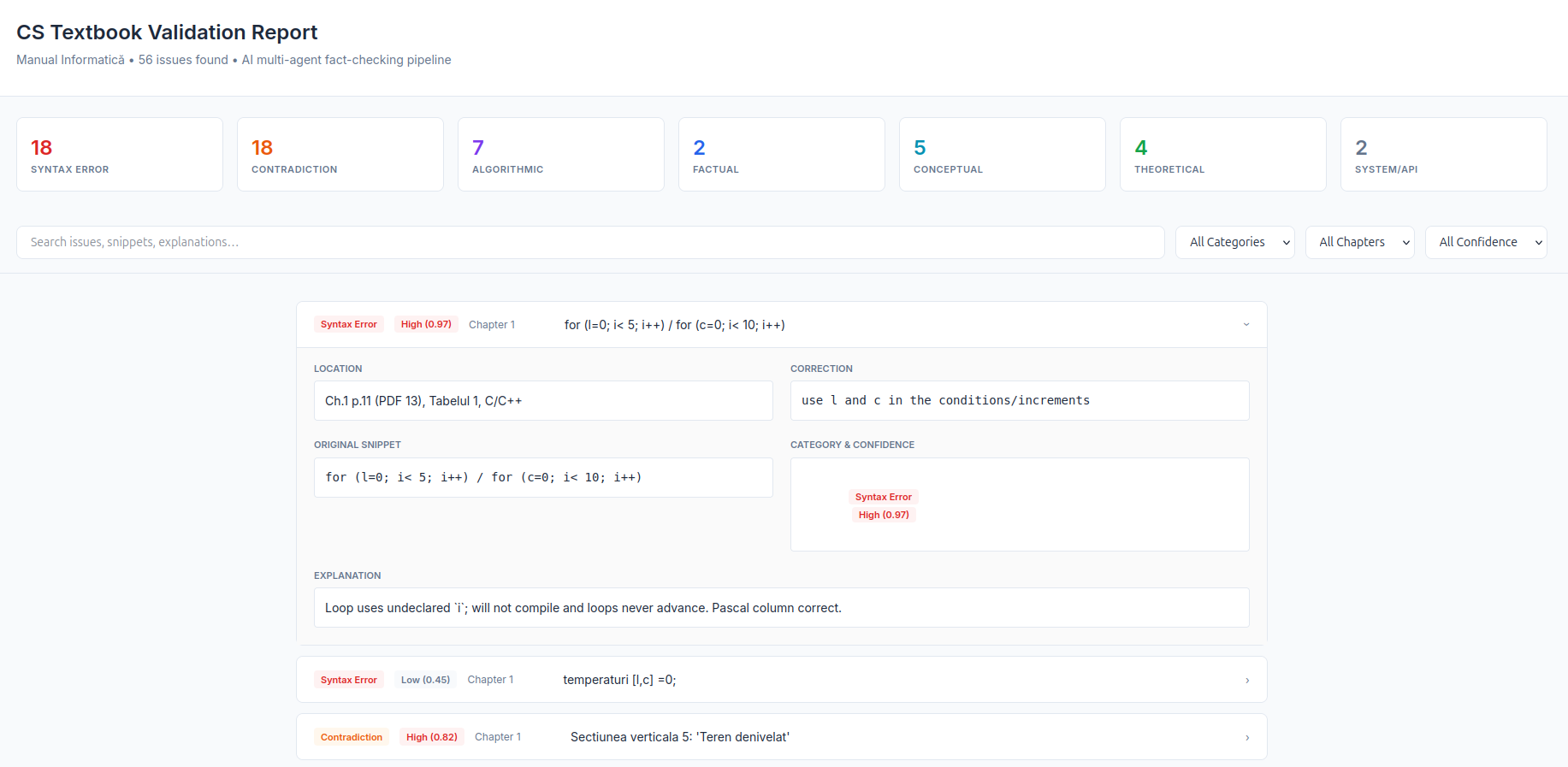}
\caption{HTML validation report for the CS upper-secondary
textbook. The statistics bar summarizes the technical findings
by category. The expanded card shows a confirmed
\textsc{Syntax} error with original fragment, suggested
correction, and explanation.}
\label{fig:report}
\end{figure}

\begin{itemize}
    \item \textbf{Syntax:} \texttt{for (l=0; i<5; i++)} --- the
    loop uses an undeclared variable \texttt{i}, so it never
    advances and does not compile (the parallel Pascal column is
    correct).
    \item \textbf{Contradiction:} in a binary-search example both
    output branches report that the searched value
    \emph{exists in the array}; the \texttt{else} branch should
    state that it does not.
    \item \textbf{Algorithmic:} a base-conversion routine
    validates its input with the impossible condition
    \texttt{b<2 \&\& b>10}, which can never hold, defeating the
    check (a non-terminating loop in Pascal, a single useless
    pass in C++).
    \item \textbf{Theoretical:} a perfect number is defined as one
    whose \emph{prime} divisors sum to their product; the correct
    definition is the sum of \emph{proper} divisors
    ($6 = 1+2+3$), and the example holds only by coincidence.
\end{itemize}

\paragraph{Expert validation.}
To move beyond raw finding counts, a domain expert manually
reviewed all 56 technical findings, labeling each as
\emph{correct}, \emph{debatable} (context-dependent or stylistic),
or \emph{incorrect} (a clear false positive): 35 findings (62.5\%)
were confirmed correct, 13 (23.2\%) debatable, and 8 (14.3\%)
incorrect, giving a strict precision of 62.5\% and a hard
false-positive rate of 14.3\%. Reliability varied sharply by
category: all 18 \textsc{Syntax} findings were correct, whereas
contradiction-type findings were the noisiest (8 correct, 5
debatable, 5 incorrect). The model's self-reported confidence
separated correct from incorrect findings only weakly (mean 0.62
vs.\ 0.43, with substantial overlap), indicating that a confidence
threshold alone is insufficient and motivating the separate Judge
and human-validation stages.

\subsection{Humanities Domain: History and Social Sciences
Textbook}

The factual and bias agent ensemble (V2 configuration, with
Tavily web search) was run on a Romanian upper-secondary
history and social sciences textbook (A364), producing
72 candidate findings: 49 grammar, 14 bias/nuance, 4
international perspective, and 5 factual errors.
Representative findings include:

\begin{itemize}
    \item \textbf{Factual:} The names of Nobel Prize
    laureates Frederick Banting and John Macleod appear as
    ``F.G. Bauting and J. Meheod''; the Exxon Valdez oil
    tanker is rendered as ``Exon Vald''; ``Three Mile Island''
    appears as ``Three Miles Island''. These require world
    knowledge unavailable to a standard grammar checker,
    motivating the LLM-based approach.
    \item \textbf{International perspective:} The textbook's
    claim that the Yalta Declaration had no role in Soviet
    policy toward Romania was flagged as contradicting
    international historiographical consensus, with archival
    evidence cited by the agent.
    \item \textbf{Bias/Nuance:} Gender-stereotyped language
    in two passages --- one attributing ``the masculine concept
    of freedom'' to men, another framing childcare as an
    exclusively female dilemma --- flagged for reformulation.
\end{itemize}

\subsection{Discussion}
The two pilot runs illustrate how domain adaptation via custom
prompts yields qualitatively different but complementary audit
profiles. On the 283-page CS textbook (V3, vision-native), the
technical agent produced 56 reviewed findings, dominated by
Syntax (18), Contradiction (18), and Algorithmic (7) errors, with
an expert-validated precision of 62.5\%.
On the 131-page history and social sciences textbook (V2, text
extraction with web search), the factual and bias agents produced
72 findings: 49 grammar, 14 bias/nuance, 4 international
perspective, and 5 factual errors. The CS run surfaces verifiable
technical errors in code and definitions, while the humanities run
surfaces factual inaccuracies, historiographical bias, and
linguistic errors that co-occur in published educational content.
All findings are candidate issues pending human expert validation.



\paragraph{Limitations and future work.}
We report a first precision estimate on the CS technical track, but the evaluation remains
preliminary: we do not yet measure recall, have no validation of
the humanities or grammar tracks, and report no Judge Agent ablation. Because the Judge trades false positives against recall (it can discard genuine errors) and rejected findings are not currently surfaced to the reviewer, such losses are invisible. The factual track relies on parametric LLM knowledge for technical domains and web search for humanities domains; neither is infallible, and both can reproduce dominant narratives or curriculum-specific conventions, producing false positives the Judge does not fully eliminate. The demonstration covers two
Romanian textbooks, one per domain; generalization to other
languages and subjects (mathematics, natural sciences), richer
inter-agent interaction, and multi-textbook evaluation remain open.
Latency is substantial given sequential per-chapter LLM calls,
suiting offline batch auditing rather than real-time use.

\section{Conclusions.}
We have presented \textbf{AI Textbook Auditor}, a modular
multi-agent pipeline for automated quality assurance of
educational materials. The system addresses a gap in
educational NLP by combining factual and technical error
detection with a domain-adaptable prompt mechanism and a
Judge Agent for false-positive filtering. Pilot runs on a
Romanian CS upper-secondary textbook (56 expert-validated
technical findings, 62.5\% precision) and a history and social
sciences textbook (72 findings spanning factual errors, bias,
and grammar) demonstrate the system's ability to surface
candidate issues across qualitatively different domains. A
grammar analysis module provides a complementary linguistic
track. Future work includes formal recall evaluation, a Judge
Agent ablation, and extension to other STEM subjects and
additional agent types (curriculum alignment, RAG-based
standard verification).

\section{Ethical considerations.}
The system is designed as an assistive tool for human
editors, not an autonomous decision-making system. All
findings require expert validation before any editorial
action is taken; the \texttt{validated} field and the
interactive interface are specifically designed to keep
a human reviewer in the loop. The textbooks analyzed
are publicly available through the Romanian national
curriculum; no student data or unpublished content is
processed. 
candidate findings, not confirmed errors --- acting on
unreviewed findings risks introducing incorrect editorial
changes to otherwise correct content.

\section*{Acknowledgments}
{\small We would like to thank Together AI for providing the API credits used for model evaluation. Additionally, the authors acknowledge the early contribution of Victor Albu to the design of the system.}

\section*{Declaration on Generative AI}
  During the preparation of this work, the authors used Claude (Anthropic), ChatGPT (OpenAI), and Gemini (Google) in order to: Grammar and spelling check, Paraphrase and reword, and Improve writing style. After using these tools and services, the authors reviewed and edited the content as needed and take full responsibility for the publication's content.

\bibliography{main}

\end{document}